\documentclass[11pt,a4paper]{article}
\usepackage[hyperref]{emnlp-ijcnlp-2019}
\usepackage{times}
\usepackage{latexsym}
\usepackage{multirow}
\usepackage{booktabs}
\usepackage{makecell}
\usepackage{natbib}
\usepackage{url}
\usepackage{algorithm2e}
\usepackage{graphicx}
\usepackage{array}
\usepackage{pgfplots}
\usepackage{tikz}
\usepackage{tikz-qtree}
\usepackage{booktabs}
\usepackage{amssymb}
\usepackage{amsthm}
\usepackage{amsmath}
\usepackage{scalerel}
\usepackage{graphicx}
\usepackage{array}
\usepackage{pgfplots}
\usepackage{tikz}
\usepackage{tikz-qtree}
\usepackage{scalerel}
\theoremstyle{definition}
\newtheorem{theorem}{Theorem}
\newtheorem{definition}[theorem]{Definition}

\theoremstyle{remark}

\newcommand{\Dom}{\emph{Dom}}
\newcommand{\Func}{\emph{Func}}
\newcommand{\leafs}{N^T_\text{leaf}}
\newcommand{\nonleafs}{N^T_\text{non-leaf}}
\newcommand{\nodes}{N^T}

\newcommand{\word}[1]{\emph{#1}}
\newcommand{\tech}[1]{\emph{#1}}
\DeclareMathOperator{\relu}{ReLU}

\definecolor{roamdarkblue}{HTML}{0499CC}
\definecolor{roamlightblue}{HTML}{03A9F4}
\definecolor{roamdarkgray}{HTML}{838A8A}
\definecolor{roamlightgray}{HTML}{B8B8B8}
\definecolor{roamgreen}{HTML}{4D8951}
\definecolor{roamblack}{HTML}{212121}
\definecolor{roamsteelblue}{HTML}{9BB8D7}
\definecolor{roamorange}{HTML}{FDBA58}
\definecolor{roamwhite}{HTML}{FAFAFA}
\definecolor{roampurple}{HTML}{876DB5}
\definecolor{mymaroon}{HTML}{881C1c}
\definecolor{hotpink}{HTML}{FF1493}
\usetikzlibrary{math}

\usepackage{url}

\aclfinalcopy % Uncomment this line for the final submission

\title{Posing Fair Generalization Tasks for Natural Language Inference}

\author{Atticus Geiger \\
  Stanford Symbolic Systems Program \\
  {\tt atticusg@stanford.edu} \\[1ex]
    \textbf{Lauri Karttunen} \\
  Stanford Linguistics \\
  {\tt laurik@stanford.edu}\\\And
  Ignacio Cases \\
  Stanford Linguistics \\
  {\tt cases@stanford.edu}\\[1ex]
  \textbf{Christopher Potts} \\
  Stanford Linguistics \\
  {\tt cgpotts@stanford.edu} \\}

\date{}

\begin{document}
\maketitle

\begin{abstract}
  Deep learning models for semantics are generally evaluated using
  naturalistic corpora. Adversarial methods, in which models are
  evaluated on new examples with known semantic properties, have begun
  to reveal that good performance at these naturalistic tasks can hide
  serious shortcomings. However, we should insist that these
  evaluations be \emph{fair} -- that the models are given data
  sufficient to support the requisite kinds of generalization. In this
  paper, we define and motivate a formal notion of fairness in this
  sense. We then apply these ideas to natural language inference by
  constructing very challenging but provably fair artificial datasets
  and showing that standard neural models fail to generalize in the
  required ways; only task-specific models that jointly compose the
  premise and hypothesis are able to achieve high performance, and
  even these models do not solve the task perfectly.
\end{abstract}

\section{Introduction}
 
Evaluations of deep learning approaches to semantics generally rely on
corpora of naturalistic examples, with quantitative metrics serving as
a proxy for the underlying capacity of the models to learn rich
meaning representations and find generalized solutions. From this
perspective, when a model achieves human-level performance on a task
according to a chosen metric, one might be tempted to say that the
task is ``solved''. However, recent adversarial testing methods, in
which models are evaluated on new examples with known semantic
properties, have begun to reveal that even these state-of-the-art
models often rely on brittle, local solutions that fail to generalize
even to examples that are similar to those they saw in training. These
findings indicate that we need a broad and deep range of evaluation
methods to fully characterize the capacities of our models.

However, for any evaluation method, we should ask 
whether it is \emph{fair}. Has the model been shown data sufficient to
support the kind of generalization we are asking of it? Unless we can
say ``yes'' with complete certainty, we can't be sure whether a failed
evaluation traces to a model limitation or a data limitation that no
model could overcome.

In this paper, we seek to address this issue by defining a formal
notion of fairness for these evaluations. The definition is quite
general and can be used to create fair evaluations for a wide range of
tasks. We apply it to Natural Language Inference (NLI) by constructing
very challenging but provably fair artificial datasets\footnote{ https://github.com/atticusg/MultiplyQuantifiedData}. We evaluate a
number of different standard architectures (variants of LSTM sequence
models with attention and tree-structured neural networks) as well as
NLI-specific tree-structured neural networks that process aligned
examples. Our central finding is that only task-specific models are
able to achieve high performance, and even these models do not solve
the task perfectly, calling into question the viability of the standard
models for semantics.

\section{Related Work}\label{sec:relatedwork}

There is a growing literature that uses targeted generalization tasks to
probe the capacity of learning models. We seek to build on this work
by developing a formal framework in which one can ask whether one of
these tasks is even possible.

% Should we have a paragraph on more informal testing out of domain?
% MultiNLI

In adversarial testing, training examples are systematically perturbed
and then used for testing. In computer vision, it is common to
adversarially train on artificially noisy examples to create a more
robust model \citep{Goodfellow-2015,Szegedy-2014}. However, in the
case of question answering, \citet{jia-2017} show that training on one
perturbation does not result in generalization to similar
perturbations, revealing a need for models with stronger
generalization capabilities. Similarly, adversarial testing has shown
that strong models for the SNLI dataset \citep{Bowman:15s} have
significant holes in their knowledge of lexical and compositional
semantics \citep{Glockner:2018,naik:2018,Nie:2018,Yanaka-etal:2019,Dasgupta:2018}. In addition, a
number of recent papers suggest that even top models exploit dataset
artifacts to achieve good quantitative results
\citep{Poliak-etal:2018,Gururangan:2018,Tsuchiya:2018}, which further
emphasizes the need to go beyond naturalistic evaluations.

Artificially generated datasets have also been used extensively to
gain analytic insights into what models are learning. These methods
have the advantage that the complexity of individual examples can be
precisely characterized without reference to the models being
evaluated. \citet{Evans:18} assess the ability of neural models to
learn propositional logic entailment. \citet{Bowman2:15} conduct
similar experiments using natural logic, % \citep{MacCartney:09},
and \citet{Veldhoen:2018} analyze models trained on those same tasks,
arguing that they fail to discover the kind of global solution we
would expect if they had truly learned natural logic. \citet{Lake:17}
apply similar methods to instruction following with an artificial
language describing a simple domain.

These methods can provide powerful insights, but the issue of fairness
looms large. For instance, \citet{Bowman:13} poses generalization
tasks in which entire reasoning patterns are held out for testing.
Similarly, \citet{Veldhoen:2018} assess a model's ability to recognize 
De Morgan's laws without any exposure to this reasoning in training.
These extremely difficult tasks break from standard evaluations in an
attempt to expose model limitations. However, these
tasks are not fair by our standards; brief formal arguments 
for these claims are given in Appendix~\ref{app:prior-unfair}.

% our standard
% deems these tasks unfair; they fail to strike a balance between
% being difficult and possible. 

%including the powerful extensions of natural logic that we explore 
%in this paper. This isn't true. Our extensions does handle demorgans lawas

\section{Compositionality and Generalization}\label{sec:compgen}

Many problems can be solved by recursively composing intermediate representations with functions along a tree structure. In the case of arithmetic, the intermediate representations are numbers and the functions are operators such a plus or minus. In the case of evaluating the truth of propositional logic sentences, the intermediate representation are truth values and the functions are logical operators such as disjunction, negation, or the material conditional. We will soon see that, in the case of NLI, the intermediate representations are semantic relations between phrases and the functions are semantic operators such as quantifiers or negation. When tasked with learning some compositional problem, we intuitively would expect to be shown how every function operates on every intermediate value. Otherwise, some functions would be underdetermined. We now formalize the idea of recursive tree-structured composition and this intuitive notion of fairness.

We first define composition trees (Section~\ref{sec:comptree}) and
show how these naturally determine baseline learning models
(Section~\ref{sec:baseline-model}). These models implicitly
define a property of fairness: a train/test split is fair if the
baseline model learns the task perfectly 
(Section~\ref{sec:fairness}). This enables us to create provably 
fair NLI tasks in Section~\ref{sec:NLItask}.

%By construction, these model perform perfectly at the tasks they are designed for. 

\begin{algorithm}[tp]
\small
\DontPrintSemicolon
\SetKwProg{Fn}{function}{}{}
\SetKwFunction{index}{index}
\SetKwFunction{childrenz}{children}
\SetKwFunction{computeTree}{compose}

\KwData{A composition tree $C = (T, \Dom, \Func)$, a node $a \in \nodes$,
  and an input $x \in \mathcal{I}_C$}
\KwResult{An output from $\Dom(a)$}
\;
%\Fn{\index{$l$}}{
%	Returns the index of leaf node $l$ according\;
%    to the standard left-to-right enumeration\;
%}
%\;
%\Fn{\children{$a$}}{
%	Returns the children of node $a$ according\;
%    to the standard left-to-right enumeration\;
%}
%\;
\Fn{\computeTree{$C,a,x$}}{
	\eIf{$a \in \leafs$}{
  	$i \leftarrow \index(a,T)$\;
  	\Return $x_i$\;
  }{
  	$c_1,\dots c_m \leftarrow \childrenz(a,T)$\;
        \Return $\Func(a)($\;
        \quad$\computeTree(C, c_1,x),\dots,$\;
        \quad$\computeTree(C, c_m,x))$\;
  }
}\;
\caption{Recursive composition up a tree. This algorithm uses helper
  functions $\texttt{children}(a,T)$, which returns the left-to-right
  ordered children of node $a$, and $\texttt{index}(a,T)$, which
  returns the index of a leaf according to left-to-right ordering.}
\label{computeTree}
\end{algorithm}

\subsection{Composition Trees}\label{sec:comptree}

A composition tree describes how to recursively compose elements from
an input space up a tree structure to produce an element in an output
space. Our baseline learning model will construct a composition tree
using training data.

\begin{definition}{(Composition Tree)}
  Let $T$ be an ordered tree with nodes $\nodes = \leafs\cup\nonleafs$,
  where $\leafs$ is the set of leaf nodes and $\nonleafs$ is the
  set of non-leaf nodes for $T$. Let $\Dom$ be a map on $\nodes$ that
  assigns a set to each node, called the \emph{domain} of the node.
  Let $\Func$ be a map on $\nonleafs$ that assigns a function to each
  non-leaf node satisfying the following property: For any $a \in
  \nonleafs$ with left-to-right ordered children $c_1,\dots, c_m$, we
  have that $\Func(a): \Dom(c_1)\times\dots\times\Dom(c_m)\to\Dom(a)$.
  We refer to the tuple $C = (T, \Dom,\Func)$ as a \emph{composition tree}.
  The \emph{input space} of this composition tree is the cartesian
  product $\mathcal{I}_C = \Dom(l_1)\times\dots\times\Dom(l_k)$, where
  $l_1,\dots, l_k$ are the leaf nodes in left-to-right order, and the
  \textit{output space} is $\mathcal{O}_C = \Dom(r)$ where $r$ is the
  root node.
\end{definition}

\begin{algorithm}[tp]
\small
\DontPrintSemicolon
\SetKwProg{Fn}{function}{}{}
\SetKwFunction{index}{index}
\SetKwFunction{childrenz}{children}
\SetKwFunction{learn}{learn}
\SetKwFunction{childrenz}{children}
\SetKwFunction{memorize}{memorize}
\SetKwFunction{initialize}{initialize}

\KwData{An ordered tree $T$ and a set of training data $\mathcal{D}$
  containing pairs $(x,Y)$ where $x$ is an input and $Y$ is
  a function defined on $\nonleafs$ providing labels at
  every node of $T$.}
\KwResult{A composition tree $ (T,\Dom, \Func)$}
\;
%\Fn{\index{$l$}}{
%	Returns the index of leaf node $l$ according\;
%    to the standard left-to-right enumeration\;
%}
%\;
%\Fn{\children{$a$}}{
%	Returns the children of node $a$ according\;
%    to the standard left-to-right enumeration\;
%}
%\;
\Fn{\learn($T,\mathcal{D}$)}{
    $\Dom , \Func \leftarrow \initialize(T)$\;
	\For{$(x,Y) \in \mathcal{D}$}{
	$\Dom,\Func \leftarrow \memorize(x,Y,T,\Dom,\Func,r)$
  }
  	\Return $(T, \Dom, \Func)$\;
}\;

\Fn{\memorize(x,Y,T,Dom,Func,a)}{
	\eIf{$a \in \leafs$}{
	$i \leftarrow \index(a,T)$\;
  	$\Dom[a] \leftarrow \Dom[a] \cup \{x_i\}$\;
  	\Return \ \Dom , \Func\;
  }{
  	$\Dom[a] \leftarrow \Dom[a] \cup \{Y(a)\}$\;
  	$c_1,\dots, c_m \leftarrow \childrenz(a,T)$\;
  	$\Func[a][(Y(c_1),\dots, Y(c_m))] \leftarrow Y(a)$\;
  \For{$k \leftarrow 1\dots m $}{
	$\Dom, \Func \leftarrow \memorize(x,Y,T,\Dom,\Func,c_k)$
  }
  	\Return \ \Dom , \Func\;
  }
}\;
\caption{Given a tree and training data with labels for every node of
  the tree, this learning model constructs a composition tree.
  This algorithm uses helper functions $\texttt{children}(a,T)$ 
  and $\texttt{index}(a,T)$, as in Algorithm \ref{computeTree}, as
  well as $\texttt{initialize}(T)$, which returns $\Dom$, a dictionary
  mapping $\nodes$ to empty sets, and $\Func$, a dictionary
  mapping $\nonleafs$ to empty dictionaries.}
\label{model1}
\end{algorithm}

A composition tree $C = (T,\Dom,\Func)$ realizes a function
$F: \mathcal{I}_C \to \mathcal{O}_C$ in the following way: For any
input $x \in \mathcal{I}_C$, this function is given by
$F(x) = \texttt{compose}(C,r,x)$, where $r$ is the root node of $T$
and $\texttt{compose}$ is defined recursively in
Algorithm~\ref{computeTree}. For a given $x \in \mathcal{I}_C$ and
$a \in \nonleafs$ with children $c_1,\dots,c_m$, we say that the
element of $\Dom(c_1)\times\dots\times\Dom(c_m)$ that is input to
$\Func(a)$ during the computation of $F(x) = \texttt{compose}(C,r,x)$
is the input realized at $\Func(a)$ on $x$ and the element of
$\Dom(a)$ that is output by $\Func(a)$ is the output realized at
$\Func(a)$ on $x$. 
At a high level, $\texttt{compose}(C,a,x)$ finds the output realized
at a node $a$ by computing node $a$'s function $\Func(a)$ with the
outputs realized at node $a$'s children as inputs. This recursion
bottoms out when the components of $x$ are provided as the outputs
realized at leaf nodes.

\subsection{A Baseline Learning Model}\label{sec:baseline-model}

Algorithm~\ref{model1} is our baseline learning model. It learns a
function by constructing a composition tree. This is equivalent to
learning the function that tree realizes, as once the composition tree
is created, Algorithm~\ref{computeTree} computes the realized
function. Because this model constructs a composition tree, it has an
inductive bias to recursively compute intermediate representations up
a tree structure. At a high level, it constructs a full composition
tree when provided with the tree structure and training data that
provides a label at every node in the tree by looping through training
data inputs and memorizing the output realized at each intermediate
function for a given input. As such, any learning model we compare to
this baseline model should be provided with the outputs realized at every
node during training.

\subsection{Fairness}\label{sec:fairness}

We define a training dataset to be fair with respect to some function
$F$ if our baseline model perfectly learns the function $F$ from that
training data.
The guiding idea behind fairness is that the training data must expose
every intermediate function of a composition tree to every possible
intermediate input, allowing the baseline model to learn a global
solution:
 
\begin{definition}{(A Property Sufficient for Fairness)}
  A property of a training dataset $\mathcal{D}$ and tree $T$ that is
  sufficient for fairness with respect to a function $F$ is that there
  exists a composition tree $C = (T,\Dom, \Func)$ realizing $F$ such
  that, for any $a \in \nonleafs$ and for any input $i$ to $\Func(a)$,
  there exists $(x, Y) \in \mathcal{D}$ where $i$ is the input realized at
  $\Func(a)$ on $x$.
\end{definition}

Not all fair datasets are challenging. For example, a scenario
in which one trains and tests on the entire space of examples
will be fair.  The
role of fairness is to ensure that, when we separately define a
challenging task, it is guaranteed to be possible. We noted in
Section~\ref{sec:relatedwork} that some challenging problems in the
literature fail to meet this minimal requirement.

\newcommand{\truthvalues}{\{\text{T}, \text{F}\}}

\newcommand{\synsem}[2]{
  \begin{array}{@{} c @{}}
    \word{#1} \\ #2
  \end{array}
}

\newcommand{\posadj}{good}
\newcommand{\negadj}{trashy}

\newcommand{\Fbut}{F_{\word{but}}}
\newcommand{\Fnot}{F_{\word{not}}}
\newcommand{\Fempty}{F_{\varepsilon}}

\newcommand{\butnode}{\synsem{but}{\Fbut(\word{Adj}, S_{2}) = F_{S_{2}}}}

\newcommand{\Sset}{\{+, -\}}
\newcommand{\negset}{\{\varepsilon, \word{not}\}}
\newcommand{\goodbad}{\adjset}
\newcommand{\topnode}{C_2(\word{Adj}, \word{but}, \word{S}) = F_{\word{but}}(\word{Adj}, S)}
\newcommand{\midnode}{C_1(\word{Neg}, \word{Adj}) = F_{\word{Neg}}(\word{Adj})}
%\newcommand{\topnode}{\synsem{\Sset}{\synsem{$\uparrow$}{C_2(\word{Adj}, \word{but}, \word{S})}}}
%\newcommand{\midnode}{\synsem{\Sset}{\synsem{$\uparrow$}{C_1(\word{Neg}, \word{Adj})}}}
%A composition tree that realizes a function performing
%    binary sentiment analysis with \word{but} phrases.
%    $\Fbut: \adjset \times \Sset \to \Sset$,
%    $\Fnot(\word{\posadj}) = -$,
%    $\Fnot(\word{\negadj}) = +$, and
%    $\Fempty: \adjset \to \Sset$.
%    We use \word{Adj} as a variable over $\adjset$ and
%    \word{Neg} as a variable over $\negset$.

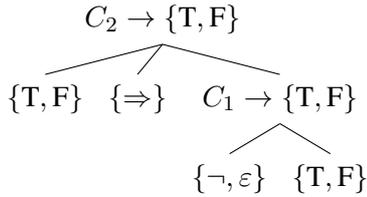
\begin{figure}[tp]
  \centering
\begin{tikzpicture}

  %\tikzset{level 1/.style={level distance=50pt}}
  %\tikzset{level 2/.style={level distance=50pt}}
  \Tree[.{$C_2 \to \truthvalues$}
  {$\truthvalues$}
  {$\{\Rightarrow\}$}
  [.{$C_1 \to \truthvalues $}
  {$\{\lnot, \varepsilon\}$}   {$\truthvalues$}  ]  ] 
\end{tikzpicture}
  \caption{A composition tree that realizes a function evaluating
  propositional sentences. We define the functions 
  $C_1(U, V_1) = U(V_1)$ and
  $C_2(V_1, \Rightarrow, V_2) = V_1 \Rightarrow V_2$ 
  where $V_1, V_2 \in \truthvalues$ and $U \in \{\lnot, \varepsilon\}$.}
  \label{proplogtree}
\end{figure}

\newcommand{\matcond}{\hspace{5pt} \Rightarrow \hspace{5pt}}

\begin{table}[tp]
  \setlength{\tabcolsep}{28pt}
  \centering
  \begin{tabular}[c]{l l}
    \toprule
    \textbf{Train} & \textbf{Test} \\
    \midrule
    T  $\matcond$ $\varepsilon$ F & T $\matcond$ $\lnot$ T \\
    T  $\matcond$  $\lnot$ F  & T $\matcond$ $\varepsilon$ T\\
    F  $\matcond$ $\lnot$ T   & F $\matcond$ $\lnot$ F \\
    F $\matcond$  $\varepsilon$ T & F $\matcond$ $\varepsilon$ F \\
    \bottomrule
  \end{tabular}  
  \caption{A fair train/test split for the evaluation problem defined
    by the composition tree in Figure~\ref{proplogtree}. We give
    just the terminal nodes; the examples are full trees.}
  \label{tab:fair-prop}
\end{table}

\subsection{Fair Tasks for Propositional Evaluation}

As a simple illustration of the above concepts, we consider the
task of evaluating the truth of a sentence from propositional 
logic. We use the standard logical operators material conditional,
$\Rightarrow$, and negation, $\lnot$, as 
well as the unary operator $\varepsilon$, which we define to be the
identity function on $\truthvalues$. We consider a small set of eight propositional sentences, all which can be seen in Table~\ref{tab:fair-prop}.  We illustrate a 
composition tree that realizes a function performing truth 
evaluation on these sentences in Figure~\ref{proplogtree}, where a leaf node
$l$ is labeled with its domain $\Dom(l)$ and a non-leaf node $a$ is 
labeled with $\Func(a) \to \Dom(a)$.

A dataset for this problem is fair if and only if it has two specific
properties.  First, the binary operator $\Rightarrow$ must be exposed to
all four inputs in $\truthvalues \times \truthvalues$ during training.  
Second, the unary operators $\lnot$ and $\varepsilon$ each must be exposed
to both inputs in $\truthvalues$. Jointly, these constraints
ensure that a model will see all the possibilities for how our 
logical operators interact with their truth-value arguments. If either
constraint is not met, then there is ambiguity about which operators the model
is tasked with learning. An example fair train/test split is given in 
Table~\ref{tab:fair-prop}.

Crucial to our ability to create a fair training dataset using only
four of the eight sentences is that $\Rightarrow$ operates on 
the intermediate representation of a truth value, abstracting away
from the specific identity of its sentence arguments. Because there
are two ways to realize T and F at the intermediate node, we can
efficiently use only half of our sentences to satisfy 
our fairness property.

\newcommand{\Peverysome}{P_{\word{every}/\word{some}}}
\newcommand{\Psomeevery}{P_{\word{some}/\word{every}}}

\newcommand{\nateq}{\ensuremath{\equiv}}
\newcommand{\natind}{\ensuremath{\mathbin{\#}}}
\newcommand{\natneg}{\ensuremath{\mathbin{^{\wedge}}}}
\newcommand{\natfor}{\ensuremath{\sqsubset}}
\newcommand{\natrev}{\ensuremath{\sqsupset}}
\newcommand{\natalt}{\ensuremath{\mathbin{|}}}
\newcommand{\natcov}{\ensuremath{\mathbin{\smile}}}

\begin{table}[tp]
\small
\setlength{\tabcolsep}{4pt}
\renewcommand{\arraystretch}{1.2}
\centering
\begin{tabular}{c c c c}
\toprule
symbol & example & set theoretic definition\\
\midrule
$x \equiv y$ & couch $\equiv$ sofa & $x = y$\\
$x \sqsubset y$ & crow $\sqsubset$ bird & $x \subset y$\\
$x \sqsupset y$ & bird $\sqsupset$ crow & $x \supset y$\\
$x \natneg y$ & human \ $\natneg$ \ nonhuman & $x \cap y = \emptyset \land  x \cup y = U$\\
$x \natalt y$ & cat $\natalt$ dog & $x \cap y = \emptyset \land x \cup y \not = U$\\
$x \natcov y$ & animal $\natcov$ nonhuman & $x \cap y \not = \emptyset \land x \cup y  = U$\\
$x \natind y$ & hungry $\#$ hippo & (all other cases)\\
\bottomrule
\end{tabular}
\caption{The seven basic semantic 
    relations of \citet{MacCartney:09}: $\mathcal{B}$ = \{$\natind$ = independence, $\sqsubset$ =
    entailment, $\sqsupset$ = reverse entailment, $\natalt$ = alternation, 
    $\natcov$ = cover, $\natneg$ = negation, $\equiv$ = equivalence\}.}
\label{tab:relations}
\end{table}

\begin{figure}[tp]
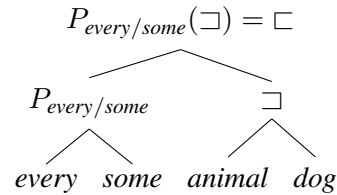

  \centering
  \Tree[.{$\Peverysome(\sqsupset) = {\sqsubset}$} 
  [.{$\Peverysome$} 
    {\word{every}} 
    {\word{some}} ] 
  [.{$\sqsupset$} {\word{animal}} {\word{dog}} ] ]
  \caption{Natural logic inference cast as composition on aligned
    semantic parse trees. The joint projectivity signature
    $\Peverysome$ operates on the semantic relation $\sqsupset$
    determined by the aligned pair \word{animal}/\word{dog}
    to determine entailment ($\sqsubset$) for the whole.
    In contrast, if we reverse \word{every} and \word{some},
    creating the example \word{some animal}/\word{every dog},
    then the joint
    projectivity signature $\Psomeevery$ operates on $\sqsupset$, which
    determines reverse entailment ($\sqsupset$).}
    \label{natlogtree}
\end{figure}

\section{Fair Artificial NLI Datasets}\label{sec:NLItask}

Our central empirical question is whether current neural models can learn to
do robust natural language inference if given fair datasets. We now present
a method for addressing this question. To do this, we need to move beyond 
the simple propositional logic example explored above, to come closer to the
true complexity of natural language. To do this, we adopt a variant of the
\emph{natural logic} developed by \citet{MacCartney:07,MacCartney:09} (see also
\citealt{valencia:91,vanBenthem:08,Icard:Moss:2013:LILT}). Natural logic
is a flexible approach to doing logical inference directly on natural 
language expressions. Thus, in this setting, we can work directly with
natural language sentences while retaining complete control over
all aspects of the generated dataset.

\subsection{Natural Logic}\label{sec:natural-logic}

We define natural logic reasoning over aligned semantic parse trees
that represent both the premise and hypothesis as a single structure
and allow us to calculate semantic relations for all phrases
compositionally. The core components are \emph{semantic relations}, which
capture the direct inferential relationships between words and phrases, and
\emph{projectivity signatures}, which encode how semantic operators
interact compositionally with their arguments. We employ the semantic relations of 
\citet{MacCartney:09}, as in Table~\ref{tab:relations}. We use $\mathcal{B}$ to denote the set containing these seven semantic relations.

The essential concept for the 
material to come is that of \emph{joint projectivity}: for a pair
of semantic functions $f$ and $g$ and a pair of inputs $X$ and $Y$
that are in relation $R$, the joint projectivity signature 
$P_{f/g}: \mathcal{B} \to \mathcal{B}$
is a function such that the relation between $f(X)$ and $g(Y)$
is $P_{f/g}(R)$. Figure~\ref{natlogtree} illustrates this with the
phrases \word{every animal} and \word{some dog}. 
%\todo{The full set of 
%joint projectivity signatures required for our datasets are 
%given in Appendix~\ref{app:projectivity}.}
We show the details of how the natural logic
of \citet{MacCartney:09}, with a small extension, determines
the joint projectivity signatures for our datasets
in Appendix~\ref{app:projectivity}.

% For example, consider the semantic functions \word{every}
% and \word{some} and a pair of inputs \word{animal} and \word{dog} that are in the 
% reverse entailment relation $\sqsupset$. The joint projectivity signature
% $P_{\word{every}/\word{some}}$ is a function telling us the relation between
% \emph{every animal} and \emph{some dog} is entailment
% $P_{every/some}(\sqsupset) = {\sqsubset}$. 

\subsection{A Fragment of Natural Language}\label{sec:frag}

\begin{figure*}
  \centering
  \input{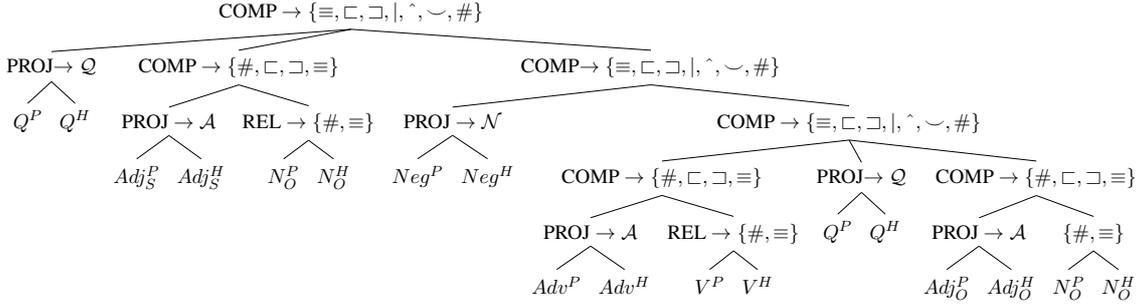}
  \caption{An aligned composition tree for inference on our set of examples $\mathcal{S}$.
    The superscripts $P$ and $H$ refer to premise and hypothesis. 
    The semantic relations are defined in Table~\ref{tab:relations}.
    The set $Q$ is $\{$\word{some}, \word{every}, \word{no}, \word{not}, \word{every}$\}$.
    The set \emph{Neg} is $\{\varepsilon, \word{not}\}$.
    $\mathcal{Q}$ is the set of 16 joint projectivity signatures between the elements of $Q$.
    $\mathcal{N}$ is the set of 4 joint projectivity signatures between $\varepsilon$ and \emph{no}.
    $\mathcal{A}$ is the set of 4 joint projectivity signatures between  $\varepsilon$ and an 
    intersective adjective or adverb.
    REL computes the semantic relations between lexical items, 
    PROJ computes the joint projectivity between two semantic functions
    (Section~\ref{sec:natural-logic} and Appendix~\ref{app:projectivity}),
    and COMP applies semantic relations to joint projectivity signatures.
    This composition tree defines over $10^{26}$ distinct examples.}
  \label{fig:bigtree}
\end{figure*}
Our fragment $G$ consists of sentences of the form:
\newcommand{\nlvar}[2]{\ensuremath{\text{#1}_{\text{#2}}}}
\begin{center}
  \nlvar{Q}{S} \nlvar{Adj}{S} \nlvar{N}{S} \nlvar{Neg}{} \nlvar{Adv}{}
  \nlvar{V}{} \nlvar{Q}{O} \nlvar{Adj}{O} \nlvar{N}{O}
  %$Q_S$ $Adj_S$ $N_S$ $Neg$ $Adv$ $V$ $Q_O$ $Adj_O$ $N_O$
\end{center}
where \nlvar{N}{S} and \nlvar{N}{O} are nouns, \nlvar{V}{} is a verb,
\nlvar{Adj}{S} and \nlvar{Adj}{O} are adjectives, and \nlvar{Adv}{} is
an adverb. \nlvar{Neg}{} is \word{does not}, and \nlvar{Q}{S} and
\nlvar{Q}{O} can be \word{every}, \word{not every}, \word{some}, or
\word{no}; in each of the remaining categories, there are 100 words.
Additionally, \nlvar{Adj}{S}, \nlvar{Adj}{O}, \nlvar{Adv}{}, and
\nlvar{Neg}{} can be the empty string $\varepsilon$, which is represented in the data
by a unique token. Semantic scope is fixed by surface order, 
with earlier elements scoping over later ones.

For NLI, we define the set of premise--hypothesis pairs
$\mathcal{S} \subset G \times G$ such that
$(s_{p}, s_{h}) \in \mathcal{S}$ iff the non-identical non-empty
nouns, adjectives, verbs, and adverbs with identical positions in
$s_{p}$ and $s_{h}$ are in the $\#$ relation.  This constraint on
$\mathcal{S}$ trivializes the task of determining the lexical
relations between adjectives, nouns, adverbs, and verbs, since the
relation is $\equiv$ where the two aligned elements are identical and
otherwise $\#$. Furthermore, it follows that distinguishing
contradictions from entailments is trivial. The only sources of
contradictions are negation and the negative quantifiers \word{no} and
\word{not every}. Consider $(s_{p}, s_{h}) \in \mathcal{S}$ and let
$C$ be the number of times negation or a negative quantifier occurs in
$s_{p}$ and $s_{h}$. If $s_{p}$ contradicts $s_{h}$, then $C$ is odd;
if $s_{p}$ entails $s_h$, then $C$ is even.

We constrain the open-domain vocabulary to stress models with learning
interactions between logically complex function words; we trivialize
the task of lexical semantics to isolate the task of compositional
semantics. We also do not have multiple morphological forms, use
artificial tokens that do not correspond to English words, and collapse
\textit{do not} and \textit{not every} to single tokens to further 
simplify the task and isolate a model's ability to perform compositional
logical reasoning.

Our corpora use the three-way labeling scheme of \tech{entailment},
\tech{contradiction}, and \tech{neutral}. To assign these labels, we
translate each premise--hypothesis pair into first-order logic and use
Prover9 \cite{McCune:2005}. We assume no expression is empty or
universal and encode these assumptions as additional premises. This
label generation process implicitly assumes the relation between
unequal nouns, verbs, adjectives, and adverbs is independence.

When we generate training data for NLI corpora from some subset
$\mathcal{S}_{\emph{train}} \subset \mathcal{S}$, we perform the following balancing.
For a given example, every adjective--noun and adverb--verb pair across the
premise and hypothesis is equally likely to have the relation
$\equiv$, $\sqsubset$, $\sqsupset$, or $\#$. Without this balancing,
any given adjective--noun and adverb--verb pair across the premise and
hypothesis has more than a 99\% chance of being in the independence
relation for values of $\mathcal{S}_{\emph{train}}$ we consider. 
Even with this step, 98\% of the sentence pairs are neutral,
so we again sample to create corpora that are balanced across the
three NLI labels. This balancing across our three NLI labels
justifies our use of an accuracy metric rather than an F1 score.

\subsection{Composition Trees for NLI}\label{sec:data}

We provide a composition tree for inference on $\mathcal{S}$ in Figure~\ref{fig:bigtree}.
This is an \emph{aligned} composition tree, as in Figure~\ref{natlogtree}: it
jointly composes lexical items from the premise and hypothesis. The leaf nodes come
in sibling pairs where one sibling is a lexical item from the premise and 
the other is a lexical item from the hypothesis. If both leaf nodes
in a sibling pair have domains containing lexical items that are semantic functions, then
their parent node domain contains the joint projectivity signatures between those semantic functions. Otherwise
the parent node domain contains the semantic relations between the lexical items
in the two sibling node domains. The root captures the overall semantic relation
between the premise  and the hypothesis, while
the remaining non-leaf nodes represent intermediate phrasal relations.

The sets $\emph{Adj}_{S}$, $\emph{N}_{S}$, $\emph{Adj}_{O}$, $\emph{N}_{O}$, 
\emph{Adv}, and \emph{V}
each have 100 of their respective open class lexical items with \emph{Adv}, 
$\emph{Adj}_{S}$, and $\emph{Adj}_{O}$ also containing the empty string $\varepsilon$.
The set $Q$ is \{\word{some}, \word{every}, \word{no}, \word{not},
\word{every}\} and the set \word{Neg} is $\{\varepsilon, \word{not}\}$.
$\mathcal{Q}$ is the set of 16 joint projectivity signatures between
the quantifiers \textit{some}, \word{every}, \word{no}, and \word{not
  every}, $\mathcal{N}$ is the set of 4 joint projectivity signatures
between the empty string $\varepsilon$ and \emph{no}, and $\mathcal{A}$ is the
set of 4 projectivity signatures between $\varepsilon$ and an
intersective adjective or adverb. These joint projectivity 
signatures were exhaustively
determined by us by hand, using the projectivity signatures 
of negation and quantifiers provided by \citet{MacCartney:09} as well
as a small extension (details in Appendix~\ref{app:projectivity}).

The function PROJ computes the joint projectivity signature 
between two semantic functions, REL computes the semantic relation between two lexical 
items, and COMP inputs semantic relations into a joint projectivity
signature and outputs the result. We trimmed the domain of every node so 
that the function of every node is surjective. Pairs of 
subexpressions containing quantifiers can be in any of the seven basic 
semantic relations; even with the contributions of open-class lexical 
items trivialized, the level of complexity remains high, and all of it 
emerges from semantic composition, rather than from lexical relations.

\subsection{A Difficult But Fair NLI Task}\label{sec:diffNLI}

A fair training dataset exposes each local function to all possible inputs.  
Thus, a fair training dataset for NLI will have the following properties. 
First, all lexical semantic relations must be included in the training
data, else the lexical targets could be underdetermined. 
Second, for any aligned semantic functions $f$ and $g$
with unknown joint projectivity signature $P_{f/g}$, and for any
semantic relation $R$, there is some training example
where  $P_{f/g}$ is exposed to the semantic
relation $R$. This ensures that the model has enough information to
learn full joint projectivity signatures. Even with these constraints in
place, the composition tree of Section~\ref{sec:comptree} determines
an enormous number of very challenging train/test splits. 
Appendix~\ref{app:datagen} fully defines the procedure for data generation.

We also experimentally verify that our baseline 
learns a perfect solution from the data we generate. The training 
set contains 500,000 examples randomly sampled from 
$\mathcal{S}_{\emph{train}}$ and the test and development sets each 
contain 10,000 distinct examples randomly sampled from 
$\bar{\mathcal{S}}_{\emph{train}}$. All random sampling is balanced 
across adjective--noun and adverb--verb relations as well as 
across the three NLI labels, as described in Section~\ref{sec:frag}. 

%\footnote{Our dataset generation code: %\url{https://github.com/atticusg/MultiplyQuantifiedData}}

\section{Models}\label{sec:models}

% \begin{figure*}
% \small
% \centering
% \begin{tabular}{|c  |c |c |c|}
% \hline
% sentence &   verb phrase & modifier phrase & single words\\
% \hline
% Every tall kid  $\varepsilon$ happily kicks every $\varepsilon$ rock  & happily kicks every $\varepsilon$ rock & happily kicks & tall $\equiv$ tall\\
% entails & $\sqsubset$ & $\sqsubset$ &kid $\equiv$ kid \\
%  No tall kid does not $\varepsilon$ kicks some large rock & $\varepsilon$ kicks some large rock & $\varepsilon$ kicks & happily $\sqsubset$ $\varepsilon$\\
%  \hline
%  negation phrase &  modifier phrase & modifier phrase & single words\\
% \hline
% $\varepsilon$ happily kicks every $\varepsilon$ rock  & tall kid & $\varepsilon$ rock & kicks $\equiv$ kicks\\
%  $|$ & $\equiv$ & $\sqsupset$ & $\varepsilon$ $\sqsupset$ large\\
% does not $\varepsilon$ kicks some large rock & tall kid & large rock & rock $\equiv$ rock\\
% \hline

% \end{tabular}
%   \caption{For any example sentence pair (top left) the neural models are trained using the a weighted sum of the error on 12 prediction tasks shown above.}
%     \label{intermediateexample}
% \end{figure*}

We consider six different model architectures:
\begin{description}\setlength{\itemsep}{0pt}

\item[CBoW] Premise and hypothesis are represented by the average of
  their respective word embeddings (continuous bag of words).

\item[LSTM Encoder] Premise and hypothesis are processed as
  sequences of words using a recurrent neural network (RNN) with LSTM
  cells, and the final hidden state of each
  serves as its representation \cite{Hochreiter:97, Elman:1990,Bowman:15s}.

\item[TreeNN] Premise and hypothesis are processed as trees,
  and the semantic composition function is a single-layer 
  feed-forward network \cite{Socher:2011,Socher2:2011}. The value of
  the root node is the semantic representation in each case.

\item[Attention LSTM] An LSTM RNN 
  with word-by-word attention \cite{Rock:15}.

\item[CompTreeNN] Premise and hypothesis are processed as a
  single aligned tree, following the structure of the composition
  tree in Figure~\ref{fig:bigtree}. 
  The semantic composition function is a single-layer feed-forward 
  network \cite{Socher:2011,Socher2:2011}. The value of
  the root node is the semantic representation of the premise and
  hypothesis together.
  
\item[CompTreeNTN] Identical to the CompTreeNN, but with a
   neural tensor network as the composition function \cite{socher:13}. 

\end{description}

For the first three models, the premise and hypothesis representations
are concatenated. For the CompTreeNN, CompTreeNTN, and Attention LSTM,
there is just a single representation of the pair. In all cases, the
premise--hypothesis representation is fed through two hidden layers and
a softmax layer. 
% Details on our procedure for hyperparameter selection
% and optimization are in Appendix~\ref{app:optimization}.

\begin{figure}
\scriptsize
\begin{tabular}{|c  |c |}
\hline
sentence &  adverb-verb phrase  \\
\hline
every tall kid  $\epsilon$ happily kicks every $\epsilon$ rock  &  happily kicks\\
\textit{entailment} & $\sqsubset$ \\
 no tall kid does not $\epsilon$ kicks some large rock & $\epsilon$ kicks \\
 \hline
 negated verb phrases &  adjective-noun phrase \\
\hline
$\epsilon$ happily kicks every $\epsilon$ rock & tall kid  \\
 $|$ & $\equiv$ \\
does not $\epsilon$ kicks some large rock & tall kid \\
\hline
verb phrases  & single words\\
 \hline
 happily kicks every $\epsilon$ rock & tall $\equiv$ tall\\
  $\sqsubset$ &kid $\equiv$ kid \\
    $\epsilon$ kicks some large rock & happily $\sqsubset$ $\epsilon$\\
   \hline
    adjective-noun phrase & single words\\
    \hline
 $\epsilon$ rock & kicks $\equiv$ kicks\\
  $\sqsupset$ & $\epsilon$ $\sqsupset$ large\\
  large rock & rock $\equiv$ rock\\
  \hline

\end{tabular}
  \caption{ For any example sentence pair (top left) the neural models are trained using the a weighted sum of the error on 12 prediction tasks shown above. The 12 errors are weighted to regularize the loss according to the length of the expressions being predicted on.}
    \label{intermediateexample}
\end{figure}

All models are initialized with random 100-dimensional word vectors
and optimized using Adam \cite{Kingma:14}. It would not be possible to use pretrained
word vectors, due to the artificial nature of our dataset. A grid
hyperparameter search was run over dropout values of $\{0,0.1,0.2,0.3\}$
on the output and keep layers of LSTM cells, learning rates of
$\{1\mathrm{e}{-2},3\mathrm{e}{-3},1\mathrm{e}{-3}, 3\mathrm{e}{-4}\}$,
L2 regularization values of
$\{0,1\mathrm{e}{-4},1\mathrm{e}{-3}, 1\mathrm{e}{-2}\}$ on all weights,
and activation functions $\relu$ and $\tanh$. Each hyperparameter
setting was run for three epochs and parameters with the highest
development set score were used for the complete training runs.

The training datasets for this generalization task are only fair if the outputs realized at every non-leaf node are provided during training just as they are in our baseline learning model. For our neural models, we accomplish this by predicting semantic relations for every subexpression pair in the scope of a node in the tree in Figure \ref{fig:bigtree} and summing the loss of the predictions together. We do not do this for the nodes labeled PROJ $\rightarrow \mathcal{Q}$ or PROJ $\rightarrow \mathcal{N}$, as the function PROJ is a bijection at these nodes and no intermediate representations are created. For any example sentence pair the neural models are trained using the a weighted sum of the error on 12 prediction tasks shown in Figure \ref{intermediateexample}. The 12 errors are weighted to regularize the loss according to the length of the expressions being predicted on.

The CompTreeNN and CompTreeNTN models are structured to create intermediate representations of these 11 aligned phrases and so intermediate predictions are implemented as in the sentiment models of \citet{socher:13}. The other models process each of the 11 pairs of aligned phrases separately. Different softmax layers are used depending on the number of classes, but otherwise the networks have identical parameters for all predictions.

% For any example sentence pair the neural models are trained using the a weighted sum of the error on 12 prediction tasks shown in Figure \ref{intermediateexample}. The 12 errors are weighted to regularize the loss according to the length of the expressions being predicted on.

\begin{table*}[tp]
  \centering
  \setlength{\tabcolsep}{20pt}
\begin{tabular}{r c c c}
\toprule
Model &  Train & Dev & Test\\
\midrule
CBoW& $88.04\pm0.68$  & $54.18\pm0.17$  & $53.99\pm0.27$ \\
%\addlinespace[0.2cm]
TreeNN& $67.01\pm12.71$  & $54.01\pm8.40$  & $53.73\pm8.36$ \\
%\addlinespace[0.2cm]
LSTM encoder& $98.43\pm0.41$  & $53.14\pm2.45$  & $52.51\pm2.78$ \\
%\addlinespace[0.2cm]
Attention LSTM& $73.66\pm9.97$  & $47.52\pm0.43$  & $47.28\pm0.95$ \\
%\addlinespace[0.2cm]
CompTreeNN& $99.65\pm0.42$  & $80.17\pm7.53$  & $80.21\pm7.71$ \\
%\addlinespace[0.2cm]
CompTreeNTN& $99.92\pm0.08$  & $90.45\pm2.48$  & $90.32\pm2.71$ \\
%\addlinespace[0.2cm]
\bottomrule
\end{tabular}

  \caption{Mean accuracy of 5 runs on our difficult but fair generalization task, with standard 95\% confidence intervals. These models are trained on the intermediate predictions described in Section \ref{sec:models}.}
  \label{tab:natlogresults}
\end{table*}

\section{Results and Analysis}\label{sec:results}

Table~\ref{tab:natlogresults} summarizes our findings on the hardest of our fair 
generalization tasks, where the training sets are minimal ones required for
fairness. The four standard neural models fail the task completely.
The CompTreeNN and CompTreeNTN, while better, are not able to solve the task perfectly
either. However, it should be noted that the CompTreeNN outperforms our four 
standard neural models by ${\approx}30\%$ and the CompTreeNTN improves on this 
by another ${\approx}10\%$. This increase in performance leads us to believe there 
may be some other composition function that solves this task perfectly.

Both the CompTreeNN and CompTreeNTN have large 95\% confidence intervals, 
indicating that the models are volatile and sensitive to random initialization.
The TreeNN also has a large $95\%$ interval. On one of the five runs, 
the TreeNN achieved a test accuracy of $65.76\%$, much higher than usual, 
indicating that this model may have more potential than the other three. 

\begin{figure*}[tp]
  \centering
  \input{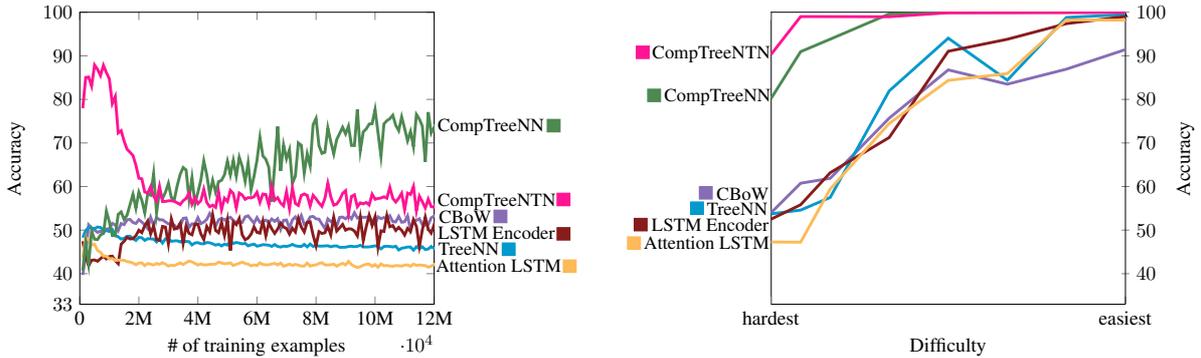}
  \caption{\textbf{Left}: Model performance on our difficult but fair generalization task throughout training. 
  \textbf{Right}: Mean accuracy of 5 runs as we move from true generalization tasks (`hardest') to problems
  in which the training set contains so much redundant encoding of the test set that the task is
  essentially one of memorization (`easiest'). Only the task-specific CompTreeNN and CompTreeNTN are able to
  do well on true generalization tasks. The other neural models succeed only where memorization suffices,
  and the CBoW model never succeeds because it does not encode word order.}
  \label{fig:genresultsgraph}
\end{figure*}

Figure~\ref{fig:genresultsgraph}, left panel, provides further insights into these results
by tracking dev-set performance throughout training.  It is evident here
that the standard models never get traction on the problem. The volatility
of the CompTreeNN and CompTreeNTN is also again evident. Notably, the CompTreeNN  
is the only model that doesn't peak in the first four training epochs, 
showing steady improvement throughout training.

% \begin{figure}[tp]
%   \centering
%   \input{Figures/ModelsDuringTraining2.tex}
%   \caption{Model performance on our difficult, but fair generalization task throughout training.}
%   \label{fig:natlogepochs}
% \end{figure}
%\begin{table*}[h]
%  \small
%  \centering
%  \input{Figures/GenResultsTable.tex}
%  \caption{Mean accuracy of 5 runs, with 95\% confidence intervals. Each model is trained with and without the %intermediate predictions shown in Figure \ref{intermediateexample}. This data is graphed in Figure %\ref{fig:genresultsgraph}. }
%  \label{tab:genresults}
%\end{table*}

% \begin{figure}[tp]
%   \centering
%   \input{Figures/GenGraph.tex}
%   \caption{Mean accuracy of 5 runs on as we ease the difficulty
%   of the generalization task while maintaining fairness (Details
%   in Appendix~\ref{app:datagen}).}
%   \label{fig:genresultsgraph}
% \end{figure}

We can also increase the number of training examples so that the training
data redundantly encodes the information needed for fairness. As we do this, the learning
problem becomes one of trivial memorization. Figure~\ref{fig:genresultsgraph},
right panel, tracks performance on this sequence of progressively more trivial problems. 
The CompTreeNN and CompTreeNTN both rapidly ascend to perfect performance. In contrast, 
the four standard models continue to have largely undistinguished performance 
for all but the most trivial problems. Finally, CBoW, while competitive with other 
neural models initially, falls behind  in a permanent way; 
its inability to account for word order prevents it from even memorizing the 
training data.

The results in Figure~\ref{fig:genresultsgraph} are for models trained to predict the semantic relations for every subexpression pair in the scope of a node in the tree in Figure~\ref{fig:bigtree} (as discussed in Section~\ref{sec:models}), but we also trained the models without intermediate predictions to quantify their impact.

All models fail on our difficult generalization task when these intermediate values are withheld. Without intermediate values this task is unfair by our standards, so this to be expected. In the hardest generalization setting the CBoW model is the only one of the four standard models to show statistically significant improvement when intermediate predictions are made. We hypothesize that the model is learning relations between open-class lexical items, which are more easily accessible in its sentence representations.
As the generalization task approaches a memorization task, the four standard models benefit more and more from intermediate predictions. In the easiest generalization setting, the four standard models are unable to achieve a perfect solution without intermediate predictions, while the CompTreeNN and CompTreeNTN models achieve perfection with or without the intermediate values. \citet{Geiger-etal:2018} show that this is due to standard models being unable to learn the lexical relations between open class lexical items when not directly trained on them. Even with intermediate predictions, the standard models are only able to learn the base case of this recursive composition.

\section{The Problem is Architecture}

One might worry that these results represent a failure of model capacity. 
However, the systematic errors remain even for much larger
networks; the trends by epoch and final results are virtually identical with 
200-dimensional rather than 100-dimensions representations. 

The reason these standard neural models fail to perform natural logic reasoning is
their architecture. The CBoW, TreeNN, and LSTM Encoder models all separately bottleneck 
the premise and hypothesis sentences into two sentence vector embeddings, so the only place
interactions between the two sentences can occur is in the two hidden layers before
the softmax layer. However, the essence of natural logic reasoning is recursive composition 
up a tree structure where the premise and hypothesis are composed jointly, so this
bottleneck proves extremely problematic. The Attention LSTM model has an architecture 
that can align and combine lexical items from the premise and hypothesis, but it cannot perform this
process recursively and also fails. The CompTreeNN and CompTreeNTN have this recursive tree structure
encoded as hard alignments in their architecture, resulting in higher performance. Perhaps in future work,
a general purpose model will be developed that can learn to perform this recursive composition 
without a hard-coded aligned tree structure.

\section{Conclusion and Future Work}\label{sec:conclusion}

It is vital that we stress-test our models of semantics using methods that go beyond
standard naturalistic corpus evaluations. Recent experiments with artificial and
adversarial example generation have yielded valuable insights here already, but it is
vital that we ensure that these evaluations are fair in the sense that they provide 
our models with achievable, unambiguous learning targets. We must carefully and 
precisely navigate the border between meaningful difficulty and impossibility.
To this end, we developed a formal notion of fairness for train/test splits.

This notion of fairness allowed us to rigorously pose the question of whether specific 
NLI models can learn to do robust natural logic 
reasoning. For our standard models, the answer is no. For our 
task-specific models, which align premise and hypothesis, the answer is more nuanced;
they do not achieve perfect
performance on our task, but they do much better than standard models. This helps us trace the problem 
to the information bottleneck formed by learning separate premise and hypothesis representations. 
This bottleneck prevents the meaningful interactions between the premise and hypothesis
that are at the core of inferential reasoning with language. Our task-specific models 
are cumbersome for real-world tasks, but they do suggest that truly robust models of 
semantics will require much more compositional interaction than is typical in today's
standard architectures.

\section*{Acknowledgments}
We thank Adam Jaffe for help developing mathematical notation and Thomas Icard for valuable discussions. This research is based in part upon work supported by the Stanford Data Science Initiative and by the NSF under Grant No. BCS-1456077. This research is based in part upon work supported by a Stanford Undergraduate Academic Research Major Grant.

\bibliographystyle{acl_natbib}
\bibliography{emnlp-ijcnlp-2019}

\clearpage
\newpage

\newcommand{\jointonly}[1]{\framebox{$#1$}}

\begin{figure*}[tp]
\begin{center}
\small
{
\renewcommand{\arraystretch}{1.4}
\begin{tabular}{l @{\hspace{36pt}} *{7}{c} @{\hspace{36pt}} *{7}{c} }
\toprule
 & \multicolumn{7}{c}{Projectivity for first argument} &\multicolumn{7}{c}{Projectivity for second argument}\\
 & $\equiv$ & $\sqsubset$ & $\sqsupset$ & $\hat{}$ & $|$ & $\smile$ & \# & $\equiv$ & $\sqsubset$ & $\sqsupset$ & $\hat{}$ & $|$ & $\smile$ & \# \\
\midrule
\word{some/every} & $\sqsupset$ & $\sqsupset$ & $\sqsupset$ & \# &\# &\# & \# & $\sqsupset$ & \# & $\sqsupset$ & $\jointonly{\natneg}$ & $\jointonly{\natalt}$ & $\jointonly{\smile}$ & \#  \\
\word{every/some} & $\sqsubset$ & $\sqsubset$ & $\sqsubset$ & \# & \# & \#& \# & $\sqsubset$ & $\sqsubset$ & \# & $\jointonly{\natneg}$ & $\jointonly{\natalt}$ & $\jointonly{\smile}$ & \#\\
\word{every/every} & $\equiv$ & $\sqsupset$ & $\sqsubset$ & $|$ & \# & $|$ & \# & $\equiv$ & $\sqsubset$ & $\sqsupset$ & $|$ & $|$ & \# & \# \\
\word{some/some} & $\equiv$ & $\sqsubset$ & $\sqsupset$ & $\smile$ & \# & $\smile$ & \# & $\equiv$ & $\sqsubset$ & $\sqsupset$ & $\smile$ & \# & $\smile$ & \# \\
\word{not/}$\varepsilon$ & $\hat{}$ & $\smile$ & $|$ & $\equiv$ & $\sqsupset$ & $\sqsubset$ & \# &- & -& -&- & -& -& -\\
$\varepsilon$\word{/not} & $\hat{}$ & $\smile$ & $|$ & $\equiv$ & $\sqsubset$ & $\sqsupset$ & \# &- & -& -&- & -& -& -\\
\word{not/not} & $\equiv$ & $\sqsupset$ & $\sqsubset$ &  $\hat{}$ & $\smile$ & $|$ & \# &- & -& -&- & -& -& -\\
 $\varepsilon/\varepsilon$& $\equiv$ & $\sqsubset$ & $\sqsupset$ & $\hat{}$ & $|$ & $\smile$ & \# &- & -& -&- & -& -& -\\
 \bottomrule
\end{tabular}

% \begin{tabular}{l | c c c c c c c | c c c c c c c }
%   \multicolumn{8}{r}{Projectivity for first argument\enspace\enspace\enspace} &\multicolumn{7}{c}{Projectivity for second argument}\\
%  & $\equiv$ & $\sqsubset$ & $\sqsupset$ & $\hat{}$ & $|$ & $\smile$ & \# & $\equiv$ & $\sqsubset$ & $\sqsupset$ & $\hat{}$ & $|$ & $\smile$ & \# \\
% \hline
% \word{some/every} & $\sqsupset$ & $\sqsupset$ & $\sqsupset$ & \# &\# &\# & \# & $\sqsupset$ & \# & $\sqsupset$ & $\hat{}$ $^*$ & $|^*$ & $\smile^*$ & \#  \\
% \word{every/some} & $\sqsubset$ & $\sqsubset$ & $\sqsubset$ & \# & \# & \#& \# & $\sqsubset$ & $\sqsubset$ & \# & $\hat{}$ $^*$ & $|^*$ & $\smile^*$ & \#\\
% \word{every/every} & $\equiv$ & $\sqsupset$ & $\sqsubset$ & $|$ & \# & $|$ & \# & $\equiv$ & $\sqsubset$ & $\sqsupset$ & $|$ & $|$ & \# & \# \\
% \word{some/some} & $\equiv$ & $\sqsubset$ & $\sqsupset$ & $\smile$ & \# & $\smile$ & \# & $\equiv$ & $\sqsubset$ & $\sqsupset$ & $\smile$ & \# & $\smile$ & \# \\
% \word{not/}$\varepsilon$ & $\hat{}$ & $\smile$ & $|$ & $\equiv$ & $\sqsupset$ & $\sqsubset$ & \# &- & -& -&- & -& -& -\\
% $\varepsilon$\word{/not} & $\hat{}$ & $\smile$ & $|$ & $\equiv$ & $\sqsubset$ & $\sqsupset$ & \# &- & -& -&- & -& -& -\\
% \word{not/not} & $\equiv$ & $\sqsupset$ & $\sqsubset$ &  $\hat{}$ & $\smile$ & $|$ & \# &- & -& -&- & -& -& -\\
%  $\varepsilon/\varepsilon$& $\equiv$ & $\sqsubset$ & $\sqsupset$ & $\hat{}$ & $|$ & $\smile$ & \# &- & -& -&- & -& -& -\\
% \end{tabular}
}
\end{center}
\setlength{\tabcolsep}{6pt}
  \caption{The four joint projectivity signatures between quantifiers \textit{every} and \word{some} and the four joint projectivity signatures between $\varepsilon$ and \word{not}. Boxed relations would not be computed in the natural logic of \citet{MacCartney:09}.}
    \label{jointprojsigs}
\end{figure*}

\appendix
\section*{Supplementary materials for `Posing Fair Generalization Tasks for Natural Language Inference'}

\section{Some Unfair Idealized Test Scenarios}\label{app:prior-unfair}

In Section~\ref{sec:diffNLI}, we explained that a fair NLI dataset must expose 
every joint projectivity signature to each semantic relation. We will now
demonstrate that the difficult generalization tasks posed by \citet{Bowman:13}
and \citet{Veldhoen:2018} are unfair in our sense. Both use the dataset provided in 
\citet{Bowman:13}, which contains examples with premises and hypotheses that
contain a single quantifier.

\citet{Bowman:13} propose the generalization tasks SUBCLASS-OUT and PAIR-OUT. 
For a particular
joint projectivity signature and semantic relation input, the SUBCLASS-OUT
generalization task holds out all examples that expose that joint
projectivity signature to that semantic relation for testing. For a particular joint
projectivity signature, the PAIR-OUT generalization task holds out all examples containing that joint projectivity signature for testing.
Both of these tasks directly violate our standard of fairness by not exposing
all joint projectivity signatures to all semantic relation inputs.

There is a certain class of examples where the premise and hypothesis 
are equivalent due to De Morgan's laws for quantifiers. 
\citet{Veldhoen:2018} propose a generalization task that holds out this 
class of examples for testing. However, these are the
only examples in which the joint projectivity signature for \word{all} and
\word{no} and the joint projectivity signature for \word{not all} and 
\word{some} are exposed to the semantic relations $\equiv$ and $\natneg$. 
This directly violates our standard of fairness by not exposing 
all joint projectivity signatures to all semantic relation inputs.

\section{Joint Projectivity}\label{app:projectivity}
The projectivity signature of a semantic function $f$ is $P_{f}: \mathcal{B} \rightarrow \mathcal{B}$ where, if the relation between $A$ and $B$ is $R$, the relation between $f(A)$ and $f(B)$ is $P_{f}(R)$. 
The natural logic of \citet{MacCartney:09} only makes use of projectivity signatures, 
and is unable to derive De Morgan's laws for quantifiers.
For example, it would label the relation between \textit{some dog eats} and \textit{every dog does not eat}
as independence $\#$ when the true semantic relation between these expressions is contradiction, $\natneg$. 
This demonstrates the need for a joint projectivity signature that directly captures such relations. 

We provide a small extension to the natural logic theory of \citet{MacCartney:09} by introducing joint projectivity signatures, which allow for new inferences involving quantifiers. The joint projectivity signature of a pair of semantic functions $f$ and $g$ is $P_{f/g}: \mathcal{B} \rightarrow \mathcal{B}$ where, if the relation between $A$ and $B$ is $R$, the relation between $f(A)$ and $g(B)$ is $P_{f/g}(R)$. It follows that the joint projectivity between a semantic function and itself is equivalent to the projectivity of that semantic function.

The joint projectivity signatures between \word{every} and \word{some} are provided in Figure~\ref{jointprojsigs} along with
the joint projectivity signatures between $\varepsilon$ and \word{not}. We mark where the natural logic
of \citet{MacCartney:09} is extended. The remaining joint projectivity signatures between the quantifiers \word{every}, \word{some}, \word{no}, and \word{not every} can be determined by composing the joint projectivity signatures of \word{every} and \word{some} with the joint projectivity signatures between \word{not} and $\epsilon$, where we parse \word{no} as \word{not some}.

\section{Fair Data Generation Algorithm}\label{app:datagen}

 We now present Algorithm \ref{model1data}, which generates fair training data of varying difficulties by restricting the way an intermediate output is realized during training based on the outputs realized by sibling nodes. The \emph{ratio} parameter determines the difficulty of the generalization task; the higher the ratio, the more permissive the training set and the easier the task.  This algorithm uses a handful of helper functions. The function $\texttt{children}(a)$ returns the left-to-right ordered children of node $a$ and the function $\texttt{cartesian\_product}(L)$ returns the Cartesian product of a 
 tuple of sets $L$. The function $\texttt{sibling\_space}(a,c_k)$ returns the set $ \Dom(c_1) \times \dots \times \Dom(c_{k-1}) \times \Dom(c_{k+1}) \times \Dom(c_m)$ where $c_1, \dots, c_m$ are the children of $a$. The function $\texttt{random\_even\_split}(S, D, \emph{ratio})$ partitions a set $S$ into $P_1$ and $P_2$ where $\frac{|P_1|}{|S|} = \emph{ratio}$ and returns a function $F: D \to \wp(S)$ ($\wp$ is the powerset function) that satisfies the follow properties: the elements in the range of $F$ are non-empty, $P_1$ is a subset of every element in the range of $F$, the union of every element in the range of $F$ is $S$, and the elements of $P_2$ are randomly and evenly distributed among the elements in the range of $F$.
 
 By our definition, a dataset is fair if it exposes every function to all possible local inputs. 
 This algorithm recursively constructs a fair training dataset. When \emph{ratio} is set to $0$ the
 dataset constructed is minimal and when \emph{ratio} is set to $1$ the dataset is the entire space
 of examples. When this algorithm is called on the root node, it returns a function mapping outputs
 to sets of inputs and the training dataset is the union of these sets. 
 
 When this algorithm is called on an intermediate node it constructs sets of partial inputs
 that expose this node's function to all possible local inputs. Then these partial inputs are recursively passed 
 to the parent node, where the process is repeated. This ensures that the function of every node is 
 exposed to every local input by the generated training data. According to the value of \emph{ratio}, we 
 constrain how local inputs are realized based on the values of their siblings. For example, the fair training
 dataset in Table~\ref{tab:fair-prop} restricts how the truth value of the right argument to $\Rightarrow$ is
 realized base on the truth value of the left argument.
 
 We experimentally verify that our training dataset is fair, though it is clearly guaranteed to be from
 the data generation algorithm.

\begin{algorithm*}[tp]
\DontPrintSemicolon
\SetKwProg{Fn}{function}{}{}
\SetKwFunction{index}{index}
\SetKwFunction{childrenz}{children}
\SetKwFunction{equivalence}{equivalence\_classes}
\SetKwFunction{random}{random\_even\_split}
\SetKwFunction{sibling}{sibling\_space}
\SetKwFunction{generate}{generate\_inputs}
\SetKwFunction{computeTree}{compose}
\SetKwFunction{cartesian}{cartesian\_product}
%\small
\KwData{A composition tree  (\emph{T},\Dom, \Func) and \emph{ratio} a number between 0 and 1 inclusive.}
\KwResult{A training dataset $\mathcal{D}$ that is fair with respect to our baseline model}
\;
%\Fn{\index{$l$}}{
%	Returns the index of leaf node $l$ according\;
%    to the standard left-to-right enumeration\;
%}
%\;
%\Fn{\children{$a$}}{
%	Returns the children of node $a$ according\;
%    to the standard left-to-right enumeration\;
%}
%\;

\Fn{\generate(T,Dom, Func, a, ratio)}{
	\eIf{$a \in \leafs$}{
	\emph{equivalence\_classes} $\leftarrow$ Dict()\;
	\For{\emph{i} $\in$ \Dom(\emph{a})}{
	\emph{equivalence\_classes[i]} $\leftarrow$ \{\emph{i}\}\;
	}
  	\Return  \emph{equivalence\_classes}\;
  }{
  $c_1,\dots c_m \leftarrow \childrenz(a)$\;
  \emph{equivalence\_classes} $\leftarrow$ Dict()\;
  $\mathcal{C} \leftarrow \Dom(c_1) \times \Dom(c_2) \dots \times \Dom(c_m)$\;
  \For{$(i_1, i_2, \dots, i_m) \in \mathcal{C}$}{
  \emph{new\_class} $\leftarrow$ \emph{List}()\\
     \For{$k \leftarrow 1 \dots m$}{
    \emph{partial\_inputs} $\leftarrow$ \generate($T,\Dom,\Func,c_k$)\;
    \emph{split} $\leftarrow$ \random(\emph{partial\_inputs}[$i_k$], \sibling(\emph{a},\emph{k}), \emph{ratio})\;
    \emph{new\_class}.\emph{append}(\emph{split}[$i_1, \dots, i_{k-1}, i_{k+1}, \dots, i_m$])\;

  }
  \emph{equivalence\_classes}[$\Func(a)(i_1,\dots,i_m)$] $\leftarrow$ \\
  \quad \emph{equivalence\_classes}[$\Func(a)(i_1,\dots,i_m)$] $\cup $ \cartesian(\emph{new\_class})\;

  }

  	\Return \emph{equivalence\_classes}\;
  }
  }\;

\caption{This model generates a training dataset that is fair with respect to our baseline model.
The output of \texttt{generate\_inputs}(\emph{T}, \Dom, \Func, \emph{a}, \emph{ratio}) is a function mapping elements of $\Dom(a)$ to sets of partial inputs that realize the element.}
\label{model1data}
\end{algorithm*}

\end{document}